\documentclass{article}

\usepackage[T1]{fontenc}
\usepackage{times}%
\usepackage{enumitem}
\usepackage{xcolor}
\usepackage{setspace}
\usepackage{float}
\usepackage[]{changebar}
\usepackage{mathtools}
\usepackage{amsmath,amssymb,amsfonts}
\usepackage{amstext}
\usepackage{graphicx}
\usepackage{stfloats}
\usepackage{authblk}

\usepackage{url}

\usepackage{graphicx, soul, color, acronym, multirow, balance, array, subfigure, url, todonotes} 

\acrodef{CNN}{Convolutional Neural Network}
\acrodef{COD}{Convolutional Object Detection}
\acrodef{ML}{Machine Learning} 
\acrodef{MSC}{Mesenchymal Stromal Cell}
\acrodef{REMPEC}{Regional Marine Pollution Emergency Response Centre for the Mediterranean Sea}
\acrodef{CMEMS}{Copernicus Marine Environmental Monitoring Service}
\acrodef{MEDESS-4MS}{Mediterranean Decision Support System for Marine Safety}
\acrodef{DaaS}{Data-as-a-Service}
\acrodef{SOA}{Service Oriented Architecture}
\acrodef{SaaS}{Software-as-a-Service}
\acrodef{AI}{Artificial Intelligence}
\acrodef{DL}{Deep Learning}
\acrodef{RAM}{Risk Analysis Model}
\acrodef{RMS}{Resource Management Service}
\acrodef{VGI}{Volunteered Geographic Information}
\acrodef{OLAP}{On-line analytical processing}
\acrodef{ICT}{Information and Communication Technologies}
\acrodef{GUI}{Graphical User Interface}
\acrodef{ENVIS}{Environmental Information System}
\acrodef{ESI}{Environmental Sensitivity Index}
\acrodef{DB}{Data Base}
\acrodef{API}{Application Programming Interface}
\acrodef{USB}{Universal Serial Bus}
\acrodef{GPS}{Global Positioning System}
\acrodef{GSM}{Global System for Mobile Communications}
\acrodef{SMS}{Short Message Service}
\acrodef{EDSS}{Environmental Decision Support System}
\acrodef{AIS}{Automatic Identification System}
\acrodef{INSPIRE}{Infrastructure for Spatial Information in the European Community}
\acrodef{GMES}{Global Monitoring for Environment and Security}
\acrodef{GIS}{Geographic Information System}
\acrodef{IT}{Information Technology}
\acrodef{VGI}{Volunteered Geographical Information}
\acrodef{AUV}{Autonomous Underwater Vehicle}
\acrodef{SAR}{Synthetic Aperture Radar}
\acrodef{MIS}{Marine Information System}
\acrodef{EMSA}{European Maritime Safety Agency}
\acrodef{PER}{Packet Erasure Rate}
\acrodef{PLR}{Packet Loss Rate}
\acrodef{BER}{Bit Error Rate}
\acrodef{RTT}{Round-Trip Time}
\acrodef{TCP}{Transmission Control Protocol}
\acrodef{AIMD}{Additive Increase Multiplicative Decrease}
\acrodef{PEP}{Performance Enhancing Proxy}
\acrodef{cwnd}{congestion window}
\acrodef{STP}{Satellite Transport Protocol}
\acrodef{BDP}{Bandwidth-Delay Product}
\acrodef{QoS}{Quality of Service}
\acrodef{LoS}{Line of Sight}
\acrodef{NLoS}{Non Line of Sight}
\acrodef{BLoS}{Beyond Line of Sight}
\acrodef{QoE}{Quality of Experience}
\acrodef{ACM}{Adaptive Coding and Modulation}
\acrodef{DAMA}{Demand Assignment Multiple Access}
\acrodef{VANET}{Vehicular Ad-Hoc Network}
\acrodef{RA}{Random Access}
\acrodef{CRA}{Contention Resolution ALOHA}
\acrodef{SA}{Slotted ALOHA}
\acrodef{DSA}{Diversity Slotted ALOHA}
\acrodef{OBU}{On-Board Unit}
\acrodef{CRDSA}{Contention Resolution Diversity Slotted ALOHA}
\acrodef{SIC}{Successive Interference Cancellation}
\acrodef{ARQ}{Automatic Repeat reQuest}
\acrodef{SC-ARQ}{Selective-Coded ARQ}
\acrodef{SR-ARQ}{Selective-Repeat ARQ}
\acrodef{IRSA}{Irregular Repetition Slotted ALOHA}
\acrodef{CGC}{Complementary Ground Component}
\acrodef{RSU}{Road Side Unit}
\acrodef{ACK}{Acknowledgment}
\acrodef{NACK}{Negative Acknowledgment}
\acrodef{DVB-SH}{Digital Video Broadcasting - Satellite Services to Handhelds}
\acrodef{DVB-H}{Digital Video Broadcasting - Handheld}
\acrodef{SACK}{Selective Acknowledgment}
\acrodef{SNACK}{Selective Negative Acknowledgment}\acrodef{SNACK}{Selective Negative Acknowledgment}
\acrodef{SNIR}{Signal to Noise plus Interference Ratio}
\acrodef{SCPS-TP}{Space Communications Protocol Specifications - Transport Protocol}
\acrodef{CCSDS}{Consultative Committee for Space Data Systems}
\acrodef{ESA}{European Space Agency}
\acrodef{NASA}{National Aeronautics and Space Administration}
\acrodef{BSM}{Broadband Satellite Multimedia}
\acrodef{RLNC}{Random Linear Network Coding}
\acrodef{NC}{Network Coding}

\title{From Human Mesenchymal Stromal Cells to Osteosarcoma Cells Classification by Deep Learning}

\author[1]{Mario D'Acunto}
\author[2]{Massimo Martinelli}
\author[2]{Davide Moroni}

\affil[1]{Institute of Biophysics, National Research Council of Italy, Via Moruzzi, 1 -- 56124-Pisa (IT), E-mail: mario.dacunto@pi.ibf.cnr.it}
\affil[2]{Institute of Information Science and Technologies, National Research Council of Italy, Via Moruzzi, 1 -- 56124-Pisa (IT), E-mail: \{massimo.martinelli, davide.moroni\}@isti.cnr.it}

\date{}

\begin{document}
\maketitle

\begin{abstract}
Early diagnosis of cancer often allows for a more vast choice of therapy opportunities. After a cancer diagnosis, staging provides essential information about the extent of disease in the body and the expected response to a particular treatment. The leading importance of classifying cancer patients at the early stage into high or low-risk groups has led many research teams, both from the biomedical and bioinformatics field, to study the application of \ac{DL} methods. The ability of \ac{DL}  to detect critical features from complex datasets is a significant achievement in early diagnosis and cell cancer progression.
In this paper, we focus the attention on osteosarcoma. Osteosarcoma is one of the primary malignant bone tumors which usually afflicts people in adolescence. Our contribution to classification of osteosarcoma cells is made as follows: a \ac{DL} approach is applied to discriminate  human \acp{MSC} from osteosarcoma cells and to classify the different cell populations under investigation. Glass slides of different cell populations were cultured including \acp{MSC}, differentiated in healthy bone cells (osteoblasts) and osteosarcoma cells, both single cell populations or mixed. Images of such samples of isolated cells (single-type of mixed) are recorded with traditional optical microscopy.  \ac{DL} is then applied to identify and classify single cells. Proper data augmentation techniques and cross-fold validation are used to appreciate the capabilities of a convolutional neural network to address the cell detection and classification problem.  Based on the results obtained on individual cells, and to the versatility and scalability of our  \ac{DL} approach, the next step will be its application to discriminate and classify healthy or cancer tissues to advance digital pathology.
\end{abstract}


\section{Introduction}
Every year, several million people die of cancer in the world due to the inaccessibility of appropriate detection schemes and consequent ineffective treatments \cite{mcguire2016world}.
Over the last decades, scientists have applied different methods to detect cancer tissues at an early stage. Such investigation is motivated by the fact that early diagnosis can facilitate the clinical management of patients. As a consequence, researchers have been examining methods for the early detection of cancers via several methods including cancer screening, solid, liquid and optical biopsy, prognostic determination, and monitoring. However, up till now, there are no known diagnostic procedures that do not hurt the physical health of patients during the process of cancer detection, being such a method invasive. Consequently,  early diagnosis should require the ability not only to identify cancer tissue as small as a single cell but having non-invasiveness as a prerequisite. 

Classification of cancer cells is hence essential research for early diagnosis and identification of differentiation and progression of cancer in a single cell \cite{Song}
\cite{Idikio}
\cite{Nahid}.

With the advent of new digital technologies in the field of medicine, \ac{AI} methods have been applied in cancer research to complex datasets in order to discover and identify patterns and relationships between them.  
\ac{ML} is a branch of \ac{AI} related to the problem of learning from data samples to the general concept of inference. In turn, \ac{DL} is a part of \ac{ML} methods based on learning data representation. \ac{DL} algorithms, in particular, convolutional networks, have rapidly become a methodology of choice for analyzing medical images. A fundamental concept in \ac{DL} is to let computers learn the features that optimally represent the data for the problem to be handled. This goal can be approached by building models (networks) composed of many layers that transform input data (in our case medical images) to outputs (e.g. a classification such as disease being present/absent) while learning increasingly higher level features.
In the last decade of application of \ac{DL} to medical images, \ac{COD} has become a successful approach to cancer analysis. 
In this paper, we have investigated the use of a \ac{COD}-based method to several differentiated samples of cells cultured on a glass slide, with the purpose to discriminate osteosarcoma cells from  \acp{MSC} (osteoblasts). 
The results are auspicious, exhibiting an accuracy of nearly one on the available dataset. These results related to the classification of cells of different malignant degree, ranging from normal to cancer cells, can generate important advantages in the study of cell seeding and cell growth. Indeed, such results allow efficient analysis of single cells simply by employing an optical microscope without using conventional biochemical methods that are time-consuming and may require a large number of cells.
The next step will be to extend the algorithm to large populations of cells and tissues with the purpose to improve digital histopathology.
The paper is organized as follows. First, related works are described in Section 2. Section 3 describes materials and methods, focusing on the procedure followed for the cell culture (\ref{sec:culture}), on the construction, augmentation, and annotation of the dataset (\ref{sec:dataset}) and, finally, on the chosen network architecture (\ref{sec:cnn:basic}). Section 4 reports the results of the training and accuracy of the method applied. Finally, Section 5 concludes the paper with discussion for future work. This paper extends our conference contribution \cite{d2018deep}.

\section{Related works}\label{sec:start}

The automatic classification of biological samples has received a lot of attention during the last years. Most of the conventional approaches rely on a feature extraction step, followed by feature classification for detecting the presence of structures of interest in biological images. Traditional methods have been based on handcrafted features, mainly consisting in descriptors of shape and appearance, including color and texture features. In this approach, general-purpose and ad hoc features are computed on the region of interest or the segmented structure of interest to gather into a single vector all the information for solving the visual task. By contrast, in \ac{DL} approaches, significant features for the visual task are not defined \emph{a priori} but they are learned during the  training process. Such a new approach has recently shown expert-level accuracy in medical image classification, improving new methods in diagnostic pathology \cite{bychkov}. Digital pathology exploits the quantification and classification of digitized tissue samples by supervised deep learning. This innovative approach to histopathology making use of digital methodologies has shown excellent results even for tasks previously considered too challenging to be accomplished with conventional image analysis methods  \cite{cirecsan2013mitosis,xie2015beyond,8036768,cnnhao,10.1007/978-3-319-59575-7_2,durr2016single}. 
In histopathology, several \ac{DL} results have recently appeared. 
In \cite{litjens2016deep}, the authors present two successful applications of \ac{DL} in reducing the workload for pathologists, namely prostate cancer identification in biopsy specimens and breast cancer metastasis detection in sentinel lymph nodes. Their work proves the potential of  \ac{DL} in increasing objectivity of diagnoses; indeed all glass slides in which prostate cancer and micro- and macro-metastases of breast cancer were present were automatically detected;  slides featuring normal tissue only could be excluded without the use of any additional immunohistochemical markers or human intervention. Similarly, in \cite{xu2015deep}, a \ac{CNN} is trained to provide a simple, efficient and effective method for achieving state-of-the-art classification and segmentation for the MICCAI 2014 Brain Tumor Digital Pathology Challenge. Transfer learning was used in their work, starting with a network pre-trained on an extensive general image database. Again, in \cite{ferreira2018classification}, the authors address the classification of breast cancer histology images using transfer learning starting with the general Inception Resnet v2 for direct labeling of the full images.
In \cite{bayramoglu2016deep}, the authors proposed two different \ac{CNN} architectures for breast cancer, namely a single task CNN is used to predict malignancy and multi-task CNN is used to predict both malignancy and image magnification level simultaneously. The results of their methods are compared using as a benchmark the BreaKHis dataset. 
All the previous works discussed above deal with general histological images to classify the whole images in order to decide whether there is or not the presence of malignant cells. Concerning the specific case of osteosarcoma, which is the focus of the present  paper, in \cite{mishra2017histopathological}, a \ac{CNN} is defined, trained and evaluated on hematoxylin and eosin stained images. The goal of their network is to assign tumor classes (viable tumor, necrosis) versus non-tumor directly to input slide images.  

Also, many tasks in digital pathology, directly or indirectly connected to tumor cell differentiation, require the classification of small clusters of cells up to a single cell, if possible.  For this purpose, differently, from the works mentioned above,  this paper investigates the classification of single cultured cells with a known grade of differentiation with a supervised DL approach.  Specifically, \ac{COD}-based DL method is applied to several differentiated samples of cells cultured on a glass slide, with the primary purpose to discriminate osteosarcoma cells from  \acp{MSC} (osteoblasts). 

Within the  ML techniques applied for the analysis of cancer cells, recently, COD has gained considerable interest \cite{d2018deep}. Besides, several methods have been proposed to address the object recognition task, and many software frameworks have been implemented to design, train and use deep learning networks (such as  Caffe \cite{jia2014caffe},  Apache MXNet \cite{mxapache} and many others). 
Among all such methods, Google TensorFlow \cite{abadi2016tensorflow} is currently one of the most used frameworks, and  its Object Detection API emerged as a potent tool for image recognition. 
Since the case study proposed in this paper requires the highest accuracy architecture allowable, we selected the Faster Region Convolutional Neural Network (Faster R-CNN) \cite{NIPS2015_5638,ren2017faster} that is an original region proposal network sharing features with the detection network that improves both region proposal quality and object detection accuracy.
Faster R-CNN uses two networks: a Region Proposal Network (RPN) to generate region proposals and a detector network to discover object instances. The RPN produces region proposals more quickly than the Selective Search \cite{Uijlings13} algorithm used in previous solutions.
By sharing information between the two networks, the accuracy is also improved, and this solution is currently the one with the best results in the latest object detection competitions.
Faster R-CNN approach can be applied using several network architecture as elemental deep features encoders. In \cite{DBLP:conf/cvpr/HuangRSZKFFWSG017} a guide for selecting the right architecture depending on speed, memory and accuracy is provided. 

Concerning general purpose \acp{COD}, evaluating a \ac{DL} approach to digital histopathology poses the problem of collecting a dataset sufficiently rich for performing an adequate training of the network. Indeed, as it is well known, \ac{DL} methods require many examples to understand and learn the best representation of an object model. Some of the works as mentioned above resorted to the use of transfer learning, starting with a network pre-trained on large datasets, such as ImageNet. However, also proper data augmentation strategies have been used with good results to overcome over-fitting issues. Conventional data augmentation methods address both the spatial and appearance domains of the images, by applying to the original images geometrical transformations (mainly orthogonal transformation such as rotations and mirroring) and/or intensity transformations (e.g. contrast stretching).  For instance, in \cite{lafarge2017domain}, the authors use spatial data augmentation (arbitrary rotation, mirroring and scaling) during the training of all models, while noticing that the most prominent source of variability in histopathology images is the staining color appearance. In \cite{wei2017deep}, they propose a so-called multi-scale fusion data augmentation method: their original database is augmented with a factor of 14 by rotation, scaling and mirroring randomly over all samples. They employed rotations by multiples of the right angle and a scale factor up to $0.8$, as well as horizontal and vertical mirroring, addressing the classification problem of breast cancer pathological images.

\section{Material and Methods}
\subsection{Cells Culture}
\label{sec:culture}
Normal, cancerous and mixed cells were cultured on glass slides. Details can be found in \cite{d2018deep}; in this paper we briefly describe the essential difference among the cell populations under investigation.
Undifferentiated MSCs were isolated from human bone marrow according to a previously reported method \cite{trombi} and used to perform three culture strategies. MSCs were plated on glass slides inside Petri dishes at a density of 20,000 cells with 10\% fetal bovine serum (FBS). The samples were cultured for 72 h, then fixed in 1\% neutral buffered formalin for 10 min at 4$^\circ$C.
Osteosarcoma cells consisted of human cells, named MG-63, were seeded on six glass slides at 10,000 cells. Finally
mixed cancer and healthy cells were plated on six glass slides inside Petri dishes at 10,000 cells with 10\% FBS.

\begin{figure}
  \centering
\includegraphics[width=0.95\columnwidth]{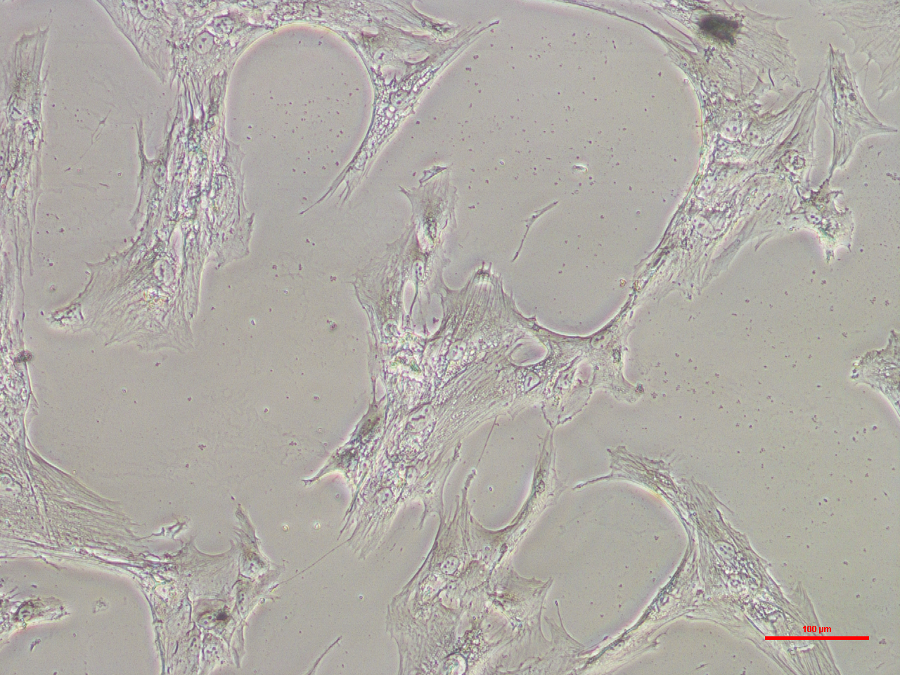}
 \includegraphics[width=0.95\columnwidth]{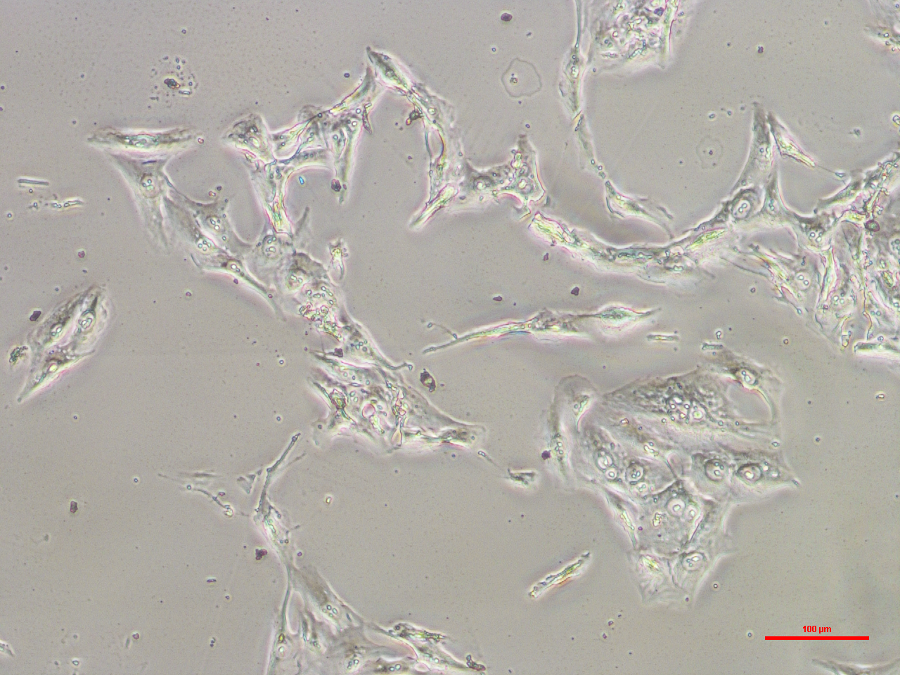}
  \caption{Morphology of osteoblast cells, (top, 10$\times$ objective, scale bar 100$\mu m$), and osteosarcoma cells (bottom, 10$\times$ objective, scale bar 100$\mu m$). }
  \label{fig:morpho}
\end{figure}

At each endpoint, all the samples were fixed in 1\% (w/v) neutral buffered formalin for 10min at 4$^\circ$C.
Morphologies are visible in Figure \ref{fig:morpho}, as imaged by an inverted microscope (Nikon Eclipse Ti-E).

\subsection{Data set collection, annotation and augmentation}
\label{sec:dataset}
A total of $N=60$ images has been collected using two different microscopes, working in two different color spaces: one acquires conventional RGB images while the other acquires monochrome images with green background density.
Experienced users have manually annotated all the images. Namely, it was requested to identify in each of the images a number of rectangular regions corresponding to particular cells and cell clusters. Five categories have been used to label the regions: 
\begin{enumerate}[label=\alph*)]
\item Single cancer cell 
\item Cancer\_cluster 
\item Single MSC cell 
\item MSC\_cluster 
\item Artifact
\end{enumerate}
 
To ease the annotation tasks, a graphical interface for performing annotation has been provided to the experts. The interface is based on the LabelImg Software \cite{tzutalin} and allows to insert multiple instances of labeled regions in each of the images in the dataset. A total of 279 objects were labeled in the images.

The dataset was therefore augmented applying both spatial and intensity transformations. With respect to other approaches that perform augmentation online directly during the training stage by applying transformations randomly, in this paper augmentation was performed offline before training. Since the dataset contains a relatively small number of images and objects when compared to large general image datasets, there is no memory and efficiency concern in the present case. For spatial transformations, we applied the dihedral group $D_4$ consisting of the symmetries of the square. Each image and the associated labeled regions were transformed accordingly, yielding a $\times8$ boost in the number of samples in the dataset. As for what  regards the color space, power law transform has been used to augment the datasets and make the results more robust with respect to illumination changes:
$$
o=c \cdot i^\gamma
$$
where $i$ represents the original input pixel value, $o$ is the output pixel value obtained after power law transformation and $c,\, \gamma$  are the parameters of the transform. In our experiments, we fixed $c=1$ and $\gamma=3/4,\, 4/5,\, 1,\, 5/4,\, 4/3$. In the case of RGB images, the power law transform was applied to each color channel. In general, such a procedure allowed for a   $\times5$ boost in dataset size.

Finally, images and labels were automatically converted into the relative TensorFlow formats. Images  were encoded into TensorFlow records, and labels were produced into Comma Separated Values (CSV) listing. Each row in the CSV listing contains the filename, the image size, the label and the top-left and bottom-right corner of the object determined by the domain expert. 



\subsection{CNN for cell detection and classification}
\label{sec:cnn:basic}
Among the possible approaches to \ac{COD}, in this paper, Faster R-CNN is adopted. Faster R-CNN  uses two sub-networks:  a deep fully convolutional network that proposes regions (named Region Proposal Network - RPN) and another module that classifies the proposed regions (classification network) \cite{NIPS2015_5638}.
The two sub-networks share the first layers which act as a feature extraction module. 
Several architectures can be used for building the feature extraction module. 
Specifically, Inception Resnet v2 model was selected in this paper and instantiated for this particular application making use of  TensorFlow  \cite{abadi2016tensorflow}. 
Transfer learning was used to cope with the limited dataset of images, which is not sufficient for dealing with training from scratch. Namely, an inference graph for Inception Resnet v2 pre-trained on COCO dataset \cite{lin2014microsoft} has been imported. 
On the basis of the feature extracted, the RPN produces candidates for regions that might contain objects of interest.  Namely, sliding a small window on the feature map, the RPN produces probabilities about the object presence in that region for region boxes of fixed aspect ratio and scale; a bounding box regressor also provides optimal size and position of the candidate rectangular areas in an intrinsic coordinate system. Candidates with a high probability of object presence are then passed to the classification network that is in charge of assessing the presence of an object category inside the region.
As a training strategy, firstly only the final fully connected layers of the two sub-networks were trained, leaving frozen all the other layers. In a fine-tuning phase, also the layers in the feature extraction module were optimized by using the training routines made available in TensorFlow.

\section{Results}
Given the limited dataset available and with the primary goal of demonstrating the applicability of \ac{DL} to the problem of cell classification, it was opted to perform $n-$fold cross validation with $n=5$ in order to obtain more statistically significant results.
The original set $A$ of $N=60$ images was partitioned into $n=5$ non-overlapping subsets $A_1$, $A_2$,$\cdots$, $A_5$ with $12$ images each. The data augmentation strategy described in Section \ref{sec:dataset} was then applied to each subset $A_i$ ($1\leq i\leq5$) producing the extended set $\bar{A}_i$ with cardinality $\#\bar{A}_i=480$ as well an associated list of labeled regions. 

Multiple training and validation sessions were then carried out. In particular for each $j$ ($1\leq j \leq5$), a network $\mathcal{N}_j$ was optimized using as training set $B_j=\bigcup_{i\neq j}\bar{A}_i$, while the set $\bar{A}_j$ might be used for validation. Notice that we opted for this partitioning approach in order to keep fully separated the training set form the validation set. Approximately, the proportion of the  split between training and validation is $4:1$, since the number of regions of interest contained in each subset $\bar{A}_j$ does not vary significantly. 

As an additional experiment, the same training procedure was repeated not taking in input the original monochrome and RGB images, but converting first all the the images to grayscale using  \cite{imagemagick}.

Each training phase lasted five days for all the training sets, using 300 regions proposals and learning parameters set to $10^{-4}$ for the first $90.000$ cycle and then reduced to $10^{-5}$. In the RPN, four scales corresponding to $1/4,\, 1/2,\, 1,\, 2$ and three aspect ratios $1/2,\, 1,\, 2$ were used.

All the inference graphs produced have been exported and tested for inference on the validation set.

\begin{figure}
	\centering
	\includegraphics[width=0.95\columnwidth]{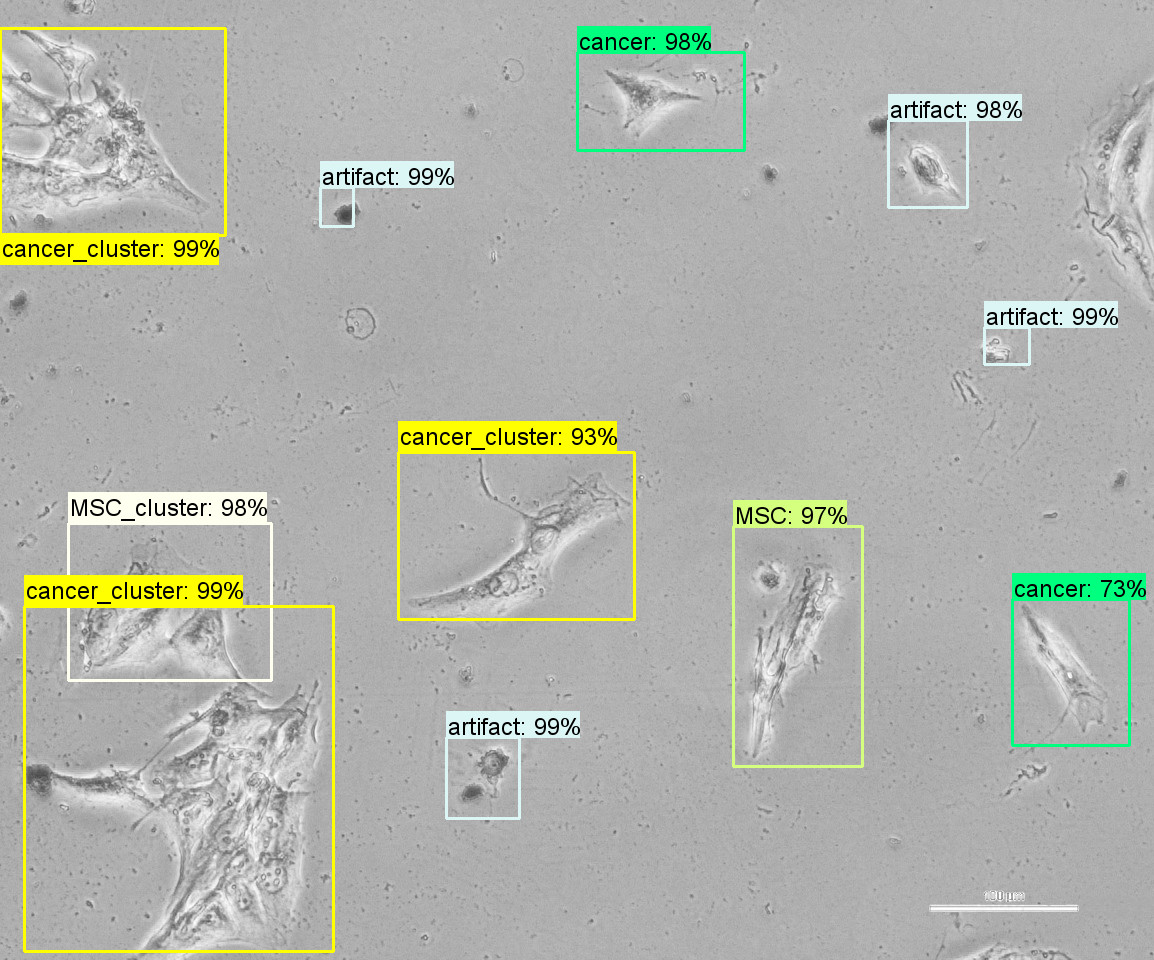}
	\caption{An example of RGB image with localized and recognized objects. Examples from the all 5 classes described in Section \ref{sec:dataset}  are shown. }
    \label{fig:train1}
\end{figure}

Figure \ref{fig:train1} reports  examples of  localization and recognition using the first graph on a RGB image.

\begin{figure}
	\centering
	\includegraphics[width=0.95\columnwidth]{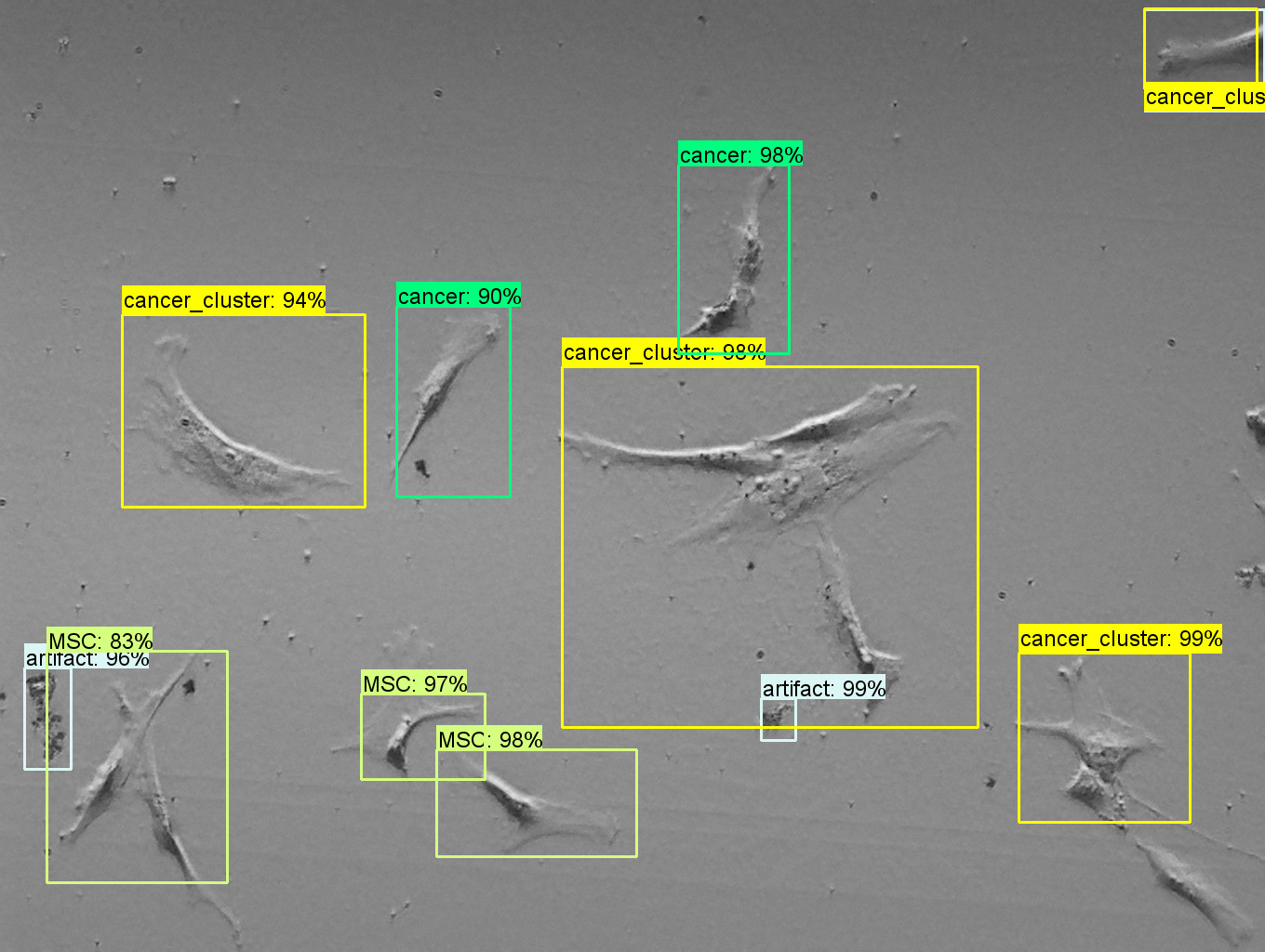}
	\caption{An example of gray-scale image with localized and recognized objects under investigation. In this case, esample from all the classes reported in Section \ref{sec:dataset} but \ac{MSC}\_cluster  are shown.}
    \label{fig:train2}
\end{figure}

\noindent Figure \ref{fig:train2} shows an example of the second graph localization and recognition on another gray-scale image.

The average accuracy obtained using RGB and the original monochrome images was $0.975\pm 0.01$. When using the images converted to grayscale very similar results have been found with an accuracy of $0.972\pm 0.005$. 
On the basis of these results, the use of color seems not to provide significant information for classification.

All training procedures have been executed  on a PC with a 4 cores 8 threads Intel(R) Core(TM) i7-4770 CPU @ 3.40 featuring 16 Giga Bytes DDR3 of RAM, an Nvidia Titan X powered with Pascal, and Ubuntu 16.04 as operating system.
Localization and recognition of new images require less than one second on a personal computer with a modern Intel I7 CPU.

\section{Conclusions}
Classification of single or small clusters of cancer cells is a crucial question for early diagnosis. In this paper, a Deep Learning approach to recognize single or small clusters of cancer cells has been presented.
The Deep Learning method adopted was based on Faster-RCNN technique and applied to several samples of cells cultured on glass slide with the purpose to discriminate osteosarcoma cells from osteo-differentiated MCSs (osteoblasts). The ability of such an algorithm to identify and classify approximately the 100\% of the investigated cells potentially will allow us to extend the method to large population cells or tissues.
These results related to the classification of cells of different malignant degree, ranging from normal to cancer cells, can have significant consequences in the study of cell seeding and cell growth. Another essential advantage of our results is that they allow efficient analysis of single cells by merely employing an optical microscope without using conventional biochemical methods that are time-consuming and may require a large number of cells.
The next step will be to extend the algorithm to large populations of cells and tissues with the purpose to improve digital histopathology.

\section*{Acknowledgements}
This research was performed in the framework of the BIO-ICT lab, a joint initiative by the Institute of Biophysics (IBF) and the Institute of Information Science and Technologies (ISTI), both of the National Research Council of Italy (CNR).

We would like to thank NVIDIA Corporation for its support: this work would have been very time-consuming without a Titan X board powered by Pascal that was granted by NVIDIA to the Signals \& Images Lab of ISTI-CNR in  2017.

In turn, we wish to thank Serena Danti, from the Department of Engineering, University of Pisa, and Luisa Trombi and Delfo D'Alessandro, from the Department of Medicine, University of Pisa, for useful support with biological samples.

\bibliographystyle{plain}

\bibliography{bibliovision}

\end{document}